\pdfoutput=1
\documentclass[
]{ceurart}

\usepackage{listings}
\usepackage{booktabs}
\usepackage{url}

\begin{document}

\copyrightyear{2026}
\copyrightclause{Copyright for this paper by its authors. Use permitted under Creative Commons License Attribution 4.0 International (CC BY 4.0).}
\conference{Posters, Demos, Blue Sky, and Tutorials at SEMANTiCS 2026, Sep 2026, Ghent, Belgium}

\title{LLM-Assisted Ontology Engineering and Construction of a French Legal Knowledge Graph}

\author[1,2]{Génesis Montenegro}[%
orcid=0009-0002-9354-2265,
]
\address[1]{Université Côte d'Azur, Inria, CNRS, I3S, Sophia-Antipolis, France}
\address[2]{Berger-Levrault, Labège, France}

\author[2]{Mokhtar Boumedyen Billami}[
orcid=0000-0003-4428-4298,
]

\author[1]{Catherine Faron}[
orcid=0000-0001-5959-5561,
]

\author[1]{Fabien Gandon}[
orcid=0000-0003-0543-1232,
]

\author[1]{Pierre Monnin}[
orcid=0000-0002-2017-8426,
]

\begin{abstract}
Maintenance regulations are complex legal texts that are difficult to exploit when addressing a specific case and challenging to integrate into operational systems. This paper presents a two-stage LLM-assisted workflow for French maintenance regulations: ontology engineering from a SEMLEG-based core ontology, followed by construction of an ontology-grounded French legal knowledge graph. The first stage consists in the open extraction of typed entities and triples from a stratified corpus sample, the normalization of labels through embedding-based fusion, and the induction of candidate object properties with their signature (domain and range). The second stage uses the resulting ontology to guide the closed extraction of triples and RDF graph construction over the full corpus. 
Experiments with \texttt{GPT-4.1} and \texttt{mistral-large-2512} show robust structured outputs, near-complete class alignment, and a substantial reduction of duplicated entities and predicates after fusion. Fewer than 20\% of triples introduce unseen properties, while lower exact signature compliance reveals new domain--range combinations for existing predicates. These results point to predicate normalization and the validation of newly observed relation signatures as key refinement steps for industrial maintenance settings.
\end{abstract}


\maketitle

\section{Introduction}

Regulatory compliance in industrial maintenance requires the interpretation of complex and evolving legal texts, making it difficult to exploit them in operational systems such as Computerized Maintenance Management Systems (CMMS). Transforming such texts into structured knowledge is therefore a key step toward queryable, reusable, and interoperable legal information. Recent work in legal information extraction and relation extraction has shown the potential of large language models (LLMs) for identifying entities and relations in legal documents~\cite{r1,r25}. However, most existing approaches remain dataset-driven and do not provide an end-to-end pipeline for moving from legal text to an ontology-grounded knowledge graph.
Ontology-based resources offer useful foundations for structuring legal and maintenance knowledge. The European Legislation Identifier (ELI) supports interoperable descriptions of legal resources, while SEMLEG models regulatory rules~\cite{r2,r3}. In the maintenance domain, existing ontologies capture activities, assets, procedures, and industrial processes~\cite{r27,r28,r29}. Nevertheless, these resources do not directly provide the domain-specific relation vocabulary required to populate a legal knowledge graph for maintenance regulations. 

To address this gap, we present an end-to-end LLM-assisted pipeline for moving from texts describing French maintenance regulations to an ontology-grounded legal knowledge graph. The workflow comprises two stages: (1) ontology engineering, where a SEMLEG-based core ontology is enriched through LLM-based extraction, embedding-based normalization, and signature-driven property induction; and (2) knowledge graph construction, where the resulting ontology guides triples extraction from the full French regulatory corpus. We contribute a compact workflow combining LLM flexibility with ontology-based structural constraints, together with a preliminary evaluation of the resulting ontology and KG.

\section{LLM-Assisted Ontology Engineering}

\paragraph{Semantic Scope and Corpus.} The semantic scope is defined through competency questions targeting maintenance obligations: involved actors, affected artifacts, contextual conditions, and legal sources. We start from SEMLEG and retain the classes required by the extraction workflow, including \textit{Actor}, \textit{Action}, \textit{Artifact}, \textit{Condition}, \textit{Source}, \textit{Location}, \textit{Reason}, \textit{Situation}, and \textit{Time}, together with the object properties whose domain and range correspond to these classes.
The corpus is built from official French legal texts retrieved from L{\'e}gifrance\footnote{L{\'e}gifrance platform: \url{https://www.legifrance.gouv.fr}.} and filtered using maintenance-oriented references from the Apave regulatory guide.\footnote{Apave regulatory guide: \url{https://france.apave.com/Actualites/Publications/Livret_Obligations_reglementaires_Apave_2026}.} These references are parsed and normalized before retrieving the corresponding legal articles and metadata, yielding a focused corpus of 6,370 regulatory articles covering 20 topics. The two stages of the workflow use this corpus differently: ontology engineering is performed on a stratified sample of 1,389 articles, corresponding to approximately 10\% of each domain--document title pair, while KG construction (ontology population) is performed over the complete corpus.

\paragraph{Open Class-Guided Triple Extraction.} Each sampled article $a_i$ is processed with three prompts. A first prompt $\pi_i^{ent}$ extracts entities from the article text $x_i$ and types them with the retained SEMLEG classes. These class declarations are injected into the prompt as a compact Turtle Light serialization, following~\cite{r15}, to keep ontology guidance readable for the LLM. 
A second prompt $\pi_i^{rel}$ then receives the article text, the extracted entities, and their classes, and generates triples $T_i$ without a predefined predicate vocabulary. Relation labels are inferred from context, while subject and object classes restrict the plausible domain--range patterns. 
Finally, a third prompt $\pi_i^{topic}$ assigns each triple to one of three topics: \textit{maintenanceActivity}, \textit{anotherLegalActivity}, or \textit{legalCrossReference}. $T_i$ is the resulting set of triples extracted from article $a_i$.
The three prompt templates are available online.\footnote{\label{footnote:github}Code and prompts: \url{https://github.com/gmontenegrou/LegiMaintLex}.}

\paragraph{Embedding-Based Fusion of Entity and Property Labels.} 
Triples classified as \textit{maintenanceActivity} are retained for ontology induction $T_{\mathrm{maint}}$; triples assigned to the other topics are discarded at this stage.
Since open extraction produces lexical variation, entity and property labels are normalized over this global set of maintenance triples.
For each class of the ontology, we consider the entities which are its instances and occur as subject or object of the extracted triples. We embed their labels and merge them when their cosine similarity is greater than $\theta_E=0.7$. The chosen canonical label is the most frequent variant. Legal references and numeric labels are excluded from this processing as they are unique or context-dependent values.  
Properties are normalized in the same way, grouping them according to the pairs of classes $(c_{e_s},c_{e_o})$ of which their subject and object are instances, and considering the same similarity threshold for merging their labels $\theta_P=0.7$~\cite{r5,r16}.
The canonical property label is selected as the most frequent variant. 
The resulting set $T_{\mathrm{maint}}^{fused}$ consists of the retained triples after replacing entity and property labels with their canonical forms.
 
\paragraph{Object Property Induction from Triples.} 
To keep object property discovery tractable while preserving diversity, we sample in  $T_{\mathrm{maint}}^{fused}$ 5\% of triples per property and ensure at least one triple per signature. The sample is split into 15 batches. For each batch, a relation-discovery prompt $\pi_k^{\mathrm{rel\_disc}}$ receives the triples, the core ontology, and the competency questions, and generates object properties with label, domain, range, textual evidence, definition, aligned question, and confidence score. These property descriptions are formalized into OWL axioms.

\paragraph{Resulting Ontology.}
Two ontology variants are generated, one using OpenAI and one using Mistral for the generation of candidate object properties. Both reuse the same SEMLEG-based core classes and properties; as we can observe in Table~\ref{tab:semlegm_variants}, their differences come from the newly induced maintenance-specific object properties. The OpenAI variant produces a broader vocabulary, with 75 maintenance-specific properties and 105 maintenance-specific signatures, while the Mistral variant is more compact, with 44 properties and 59 signatures. The two variants share 21 maintenance-related properties and 18 signatures, including \texttt{appliesTo}, \texttt{composedOf}, \texttt{performedAtLocation}, and \texttt{responsibleFor}, which capture recurring relations for applicability, composition, location, and responsibility.

\begin{table}[!ht]
\caption{Comparison of the constructed SemLegM ontology variants.}
\label{tab:semlegm_variants}
\footnotesize
\begin{tabular}{lrr}
\toprule
 & \textbf{OpenAI} & \textbf{Mistral} \\ \midrule

\# Maintenance-specific properties & 75 & 44 \\ 
\# Maintenance-specific signatures & 105 & 59 \\ \bottomrule
\end{tabular}
\end{table}

Qualitatively, OpenaAI tends to induce more fine-grained and expressive predicates, including relations for purpose, sequencing, and interaction (e.g., \texttt{aimsToAction}, \texttt{precededBy}, \texttt{transmittedTo}). Mistral tends to produce a more conservative vocabulary oriented toward normative and compliance relations (e.g., \texttt{hasModality}, \texttt{verifiedBy}, \texttt{performedUnderCondition}). We keep the variants separate to compare how different ontologies affect downstream Knowledge Graph construction.

\section{Automatic Construction of a French Legal Knowledge Graph}

The constructed ontology variants are used for guiding closed large-scale triple extraction over the full corpus. Each ontology variant is paired with the LLM that generated it. In contrast to the ontology generation, where triple generation is open, the KG generation uses signature-level prompt guidance: the triple generation prompt receives the article text, the extracted entities, and ontology-derived constraints, including admissible object properties with their signatures. These constraints guide generation at prompt level, but are not enforced through a formal validation or repair loop.

After generation, the same fusion strategy is applied to reduce entity and predicate variation. The outputs are organized into intermediate tabular representations for legal documents, metadata, triples, and entity mentions, and then lifted to RDF using declarative RML mappings. The RDF model combines established vocabularies:\footnote{Vocabulary URLs: ELI \url{http://data.europa.eu/eli/ontology\#}; DCTERMS \url{http://purl.org/dc/terms/}; CNT \url{http://www.w3.org/2011/content\#}; SKOS \url{http://www.w3.org/2004/02/skos/core\#}; PROV \url{http://www.w3.org/ns/prov\#}; Web Annotation \url{http://www.w3.org/ns/oa\#}.} ELI and DCTERMS for legal documents and metadata, CNT for textual content, SKOS and PROV for extracted entities and provenance, and Web Annotation for links between mentions and textual evidence. Relations are represented as reified statements typed as \texttt{rdf:Statement} and \texttt{semleg:ExtractedRelation}, which makes it possible to attach evidence and contextual metadata to each extracted relation.

\begin{table}[!ht]
    \caption{Statistics of the  knowledge graph variants with (F) or without (NF) fusion.}
    \label{tab:stats-kg}
    \footnotesize
    \begin{tabular}{lrrrr}
    \toprule
         & \textbf{Mistral NF} & \textbf{Mistral F} & \textbf{OpenAI NF} & \textbf{OpenAI F} \\ \midrule
       \# RDF triples & 2,119,485 & 1,131,066 & 1,507,755 & 1,311,916 \\
        \# Classes & 18 & 15 & 15 & 12 \\ 
        \# Entities & 74,035 & 20,827 & 53,804 & 38,339 \\ 
        \# Object properties & 2,643 & 500 & 2,649 & 2,031 \\
        \# Signatures & 4,694 & 1,636 & 4,374 & 3,398 \\ 
        \# \texttt{rdf:Statement} instances & 75,870 & 75,870 & 51,657 & 51,657 \\ 
        \# Annotations & 119,063 & 39,055 & 83,879 & 70,930 \\ \bottomrule
    \end{tabular}
\end{table}

Table~\ref{tab:stats-kg} summarizes the size and variability of the generated graphs. The fusion clearly preserves the extracted relation statements while substantially reducing duplicated entities, predicates, signatures, and annotations, especially for the Mistral-based graph. This shows that normalization is a central step for turning raw LLM triple generation into a more compact and usable ontology-grounded KG.

\section{Evaluation of the Ontology and Knowledge Graph}

\paragraph{Experimental Setup}
Experiments were run with a Python pipeline using OpenAI (\texttt{GPT-4.1} and \texttt{text-embedding-3-large} 
) and Mistral (\texttt{mistral-large-2512} and \texttt{mistral-embed-2312})
under the same prompting protocol. Temperature was set to 0 for deterministic generation; we used a fixed context window size of 4k tokens for class-guided entity extraction and 10k for relation discovery. We evaluate both syntactic robustness and semantic alignment of the generated relations with the constructed ontology. Datasets are available online.\footnote{Datasets: \url{https://drive.google.com/drive/folders/1GfJPdJiGv_dZuC9utOgCTyzmeYb5lYU0?usp=drive_link}.}

\paragraph{Quantitative Analysis}
We use four compact metrics: JSON validity ($R_{\text{JSON}}$), class coverage ($R_{\text{class}}$), object property coverage ($R_{\text{prop}}$), and exact relation signature compliance ($R_{\text{sig}}$). $R_{\text{prop}}$ measures whether a triple uses a property already present in the ontology, whereas $R_{\text{sig}}$ additionally measures whether the property occurs with an expected domain--range combination. Table~\ref{tab:kg_extraction_evaluation} shows robust structured output and near-complete class alignment. It also shows that fewer than 20\% of triples introduce previously unseen properties in all settings, and much less after fusion. The lower $R_{\text{sig}}$ values therefore point less to a lack of predicate vocabulary than to new or unexpected class combinations for existing predicates, such as frequent predicates reused across broader contexts than initially encoded in the ontology.

\begin{table}[ht]
\caption{Evaluation of legal relation statements with (F) or without (NF) fusion.}
\label{tab:kg_extraction_evaluation}
\footnotesize
\begin{tabular}{lrrrr}
\toprule
\textbf{Metric} & \textbf{Mistral NF} & \textbf{Mistral F} & \textbf{OpenAI NF} & \textbf{OpenAI F} \\
\midrule
$R_{\text{JSON}}$ & 100.00\% & 100.00\% & 100.00\% & 100.00\% \\
$R_{\text{class}}$ & 99.97\% & 99.97\% & 99.99\% & 99.99\% \\
$R_{\text{sig}}$ & 49.86\% & 72.61\% & 56.94\% & 61.03\% \\ 
$R_{\text{prop}}$ & 82.51\% & 96.42\% & 81.52\% & 85.11\% \\ \bottomrule
\end{tabular}
\end{table}

\paragraph{Qualitative Analysis}
We further inspect fused graphs by checking inconsistent relation signatures whose predicate suggests an expected target class (e.g., \textit{hasTime} should point to \textit{Time}). Mistral fusion yields 39 inconsistent signatures, while OpenAI fusion yields 52. Most true errors come from entity typing mistakes propagated to relations, such as legal sources classified as \textit{Artifact}, or from predicates that encode an incorrect expected object type. In general, both models encode modality or polarity directly in predicates (e.g., \texttt{cannotApplyFor}, \texttt{mustNotExceed}), which increase variation and occasionally generate French predicates despite prompts explicitly requiring English labels. These cases show that predicate fusion must balance vocabulary unification with semantic precision: over-normalizing predicates may improve compactness, but can hide distinctions that are critical in industrial maintenance, where legal obligations, prohibitions, responsibilities, and conditions may imply operational risks. Competency-question tests, implemented as SPARQL queries, additionally confirm that the graph can retrieve actor roles and legal justifications for maintenance actions.

\section{Conclusion and Future Work}
We presented an LLM-assisted pipeline for constructing an ontology-grounded French legal knowledge graph for maintenance regulations. The pipeline extracts typed entities and open relations, normalizes labels through embedding-based fusion, induces candidate object properties from relation signatures, and uses the resulting ontology to guide KG construction. 
The evaluation results show strong structured-output validity and class alignment. They also show that most triples reuse properties already present in the ontology, with fewer than 20\% involving previously unseen properties, while signature compliance reveals new domain--range combinations for existing predicates.
Future work will therefore focus on iterative ontology refinement, not only by adding missing properties, but also by validating and integrating newly observed domain--range signatures. We will also improve entity and predicate fusion and explore formal validation mechanisms, e.g., using SHACL, to detect and repair invalid signatures. Finally, we will study how the generated KG can support GraphRAG-based consultation of maintenance regulations and how it can be updated when legal provisions evolve.

\textbf{Acknowledgement.} This work has been supported by the French government, through the 3IA Côte d’Azur Investments in the project managed by the National Research Agency (ANR) with the reference number ANR-23-IACL-0001, and through the France 2030 investment plan managed by the National Research Agency (ANR), as part of the Initiative of Excellence Université Côte d’Azur under reference number ANR- 15-IDEX-01.

\section*{Declaration on Generative AI}

During the preparation of this paper the authors used GPT-5.3 and Codex-5.5 for grammar and spelling check, paraphrase and reword, and formatting assistance. After using this tool, the authors reviewed and edited the content as needed and takes full responsibility for the publication’s content.

\bibliography{biblio}

\begin{thebibliography}{10}
\expandafter\ifx\csname natexlab\endcsname\relax\def\natexlab#1{#1}\fi
\providecommand{\url}[1]{\texttt{#1}}
\providecommand{\href}[2]{#2}
\providecommand{\path}[1]{#1}
\providecommand{\DOIprefix}{doi:}
\providecommand{\ArXivprefix}{arXiv:}
\providecommand{\URLprefix}{URL: }
\providecommand{\Pubmedprefix}{pmid:}
\providecommand{\doi}[1]{\href{http://dx.doi.org/#1}{\path{#1}}}
\providecommand{\Pubmed}[1]{\href{pmid:#1}{\path{#1}}}
\providecommand{\bibinfo}[2]{#2}
\ifx\xfnm\relax \def\xfnm[#1]{\unskip,\space#1}\fi
\bibitem[{Premasiri et~al.(2025)Premasiri, Ranasinghe, Mitkov, El-Haj, and Frommholz}]{r1}
\bibinfo{author}{D.~Premasiri}, \bibinfo{author}{T.~Ranasinghe}, \bibinfo{author}{R.~Mitkov}, \bibinfo{author}{M.~El-Haj}, \bibinfo{author}{I.~Frommholz},
\newblock \bibinfo{title}{Survey on legal information extraction: Current status and open challenges},
\newblock \bibinfo{journal}{Knowledge and Information Systems} \bibinfo{volume}{67} (\bibinfo{year}{2025}) \bibinfo{pages}{11287--11358}. \DOIprefix\doi{10.1007/s10115-025-02600-5}.
\bibitem[{Li and Yi(2024)}]{r25}
\bibinfo{author}{S.~Li}, \bibinfo{author}{L.~Yi},
\newblock \bibinfo{title}{A few-shot entity relation extraction method in the legal domain based on large language models},
\newblock in: \bibinfo{booktitle}{Proceedings of the 2024 Guangdong-Hong Kong-Macao Greater Bay Area International Conference on Digital Economy and Artificial Intelligence}, \bibinfo{publisher}{ACM}, \bibinfo{address}{Hong Kong, China}, \bibinfo{year}{2024}, pp. \bibinfo{pages}{580--586}. \DOIprefix\doi{10.1145/3675417.3675513}.
\bibitem[{Breton et~al.(2022)Breton, Billami, Chevalier, and Trojahn}]{r2}
\bibinfo{author}{J.~Breton}, \bibinfo{author}{M.~B. Billami}, \bibinfo{author}{M.~Chevalier}, \bibinfo{author}{C.~Trojahn},
\newblock \bibinfo{title}{Semantic model for the legal maintenance: The case of semantic annotation of france legislative and regulatory texts},
\newblock in: \bibinfo{booktitle}{Workshop on Methodologies for Translating Legal Norms into Formal Representations (LN2FR 2022) in conjunction with JURIX 2022}, \bibinfo{address}{Saarbr{\"u}cken, Germany}, \bibinfo{year}{2022}. \URLprefix \url{https://ut3-toulouseinp.hal.science/hal-04212544}.
\bibitem[{Breton et~al.(2024)Breton, Billami, Chevalier, and Trojahn}]{r3}
\bibinfo{author}{J.~Breton}, \bibinfo{author}{M.~B. Billami}, \bibinfo{author}{M.~Chevalier}, \bibinfo{author}{C.~Trojahn},
\newblock \bibinfo{title}{Leveraging semantic model and {LLM} for bootstrapping a legal entity extraction: An industrial use case},
\newblock in: \bibinfo{booktitle}{Knowledge Graphs in the Age of Language Models and Neuro-Symbolic AI: Proceedings of the 20th International Conference on Semantic Systems, 17--19 September 2024, Amsterdam, The Netherlands}, Studies on the Semantic Web, \bibinfo{publisher}{IOS Press}, \bibinfo{address}{Amsterdam, Netherlands}, \bibinfo{year}{2024}, pp. \bibinfo{pages}{20--36}. \DOIprefix\doi{10.3233/ssw240004}.
\bibitem[{Woods et~al.(2024)Woods, Selway, Bikaun, Stumptner, and Hodkiewicz}]{r27}
\bibinfo{author}{C.~Woods}, \bibinfo{author}{M.~Selway}, \bibinfo{author}{T.~Bikaun}, \bibinfo{author}{M.~Stumptner}, \bibinfo{author}{M.~Hodkiewicz},
\newblock \bibinfo{title}{An ontology for maintenance activities and its application to data quality},
\newblock \bibinfo{journal}{Semantic Web} \bibinfo{volume}{15} (\bibinfo{year}{2024}) \bibinfo{pages}{319--352}. \DOIprefix\doi{10.3233/SW-233299}.
\bibitem[{Carriero et~al.(2025)Carriero, Scrocca, Baroni, Azzini, and Celino}]{r28}
\bibinfo{author}{V.~A. Carriero}, \bibinfo{author}{M.~Scrocca}, \bibinfo{author}{I.~Baroni}, \bibinfo{author}{A.~Azzini}, \bibinfo{author}{I.~Celino}, \bibinfo{title}{Procedural knowledge ontology ({PKO})}, \bibinfo{year}{2025}. \URLprefix \url{https://arxiv.org/abs/2503.20634v1}. \DOIprefix\doi{10.1007/978-3-031-94578-6_19}.
\bibitem[{Hodkiewicz et~al.(2024)Hodkiewicz, Woods, Selway, and Stumptner}]{r29}
\bibinfo{author}{M.~Hodkiewicz}, \bibinfo{author}{C.~Woods}, \bibinfo{author}{M.~Selway}, \bibinfo{author}{M.~Stumptner}, \bibinfo{title}{{IOF-Maint}: Modular maintenance ontology}, \bibinfo{year}{2024}. \URLprefix \url{http://arxiv.org/abs/2404.05224}. \DOIprefix\doi{10.26182/chzp-vs60}.
\bibitem[{Ringwald et~al.(2024)Ringwald, Gandon, Faron, Michel, and Akl}]{r15}
\bibinfo{author}{C.~Ringwald}, \bibinfo{author}{F.~Gandon}, \bibinfo{author}{C.~Faron}, \bibinfo{author}{F.~Michel}, \bibinfo{author}{H.~A. Akl},
\newblock \bibinfo{title}{12 shades of {RDF}: Impact of syntaxes on data extraction with language models},
\newblock in: \bibinfo{booktitle}{Lecture Notes in Computer Science}, Lecture Notes in Computer Science, \bibinfo{publisher}{Springer Nature Switzerland}, \bibinfo{address}{Hersonissos, Greece}, \bibinfo{year}{2024}, pp. \bibinfo{pages}{81--91}. \DOIprefix\doi{10.1007/978-3-031-78952-6_8}.
\bibitem[{Lairgi et~al.(2024)Lairgi, Moncla, Cazabet, Benabdeslem, and Cl{\'e}au}]{r5}
\bibinfo{author}{Y.~Lairgi}, \bibinfo{author}{L.~Moncla}, \bibinfo{author}{R.~Cazabet}, \bibinfo{author}{K.~Benabdeslem}, \bibinfo{author}{P.~Cl{\'e}au}, \bibinfo{title}{{iText2KG}: Incremental knowledge graphs construction using large language models}, \bibinfo{year}{2024}. \URLprefix \url{http://arxiv.org/abs/2409.03284}. \DOIprefix\doi{10.48550/arXiv.2409.03284}.
\bibitem[{Lairgi et~al.(2025)Lairgi, Moncla, Benabdeslem, Cazabet, and Cl{\'e}au}]{r16}
\bibinfo{author}{Y.~Lairgi}, \bibinfo{author}{L.~Moncla}, \bibinfo{author}{K.~Benabdeslem}, \bibinfo{author}{R.~Cazabet}, \bibinfo{author}{P.~Cl{\'e}au}, \bibinfo{title}{{ATOM}: Adaptive and optimized dynamic temporal knowledge graph construction using {LLMs}}, \bibinfo{year}{2025}. \URLprefix \url{http://arxiv.org/abs/2510.22590}. \DOIprefix\doi{10.48550/arXiv.2510.22590}.

\end{thebibliography}

\end{document}